\documentclass[10pt,twocolumn,letterpaper]{article}

\usepackage{bm}
\usepackage{cvpr}
\usepackage{times}
\usepackage{epsfig}
\usepackage{graphicx}
\usepackage{amsmath}
\usepackage{amssymb}

\usepackage{multirow}
\usepackage{pifont}
\usepackage{tabularx}
\usepackage{subfigure}

\newcolumntype{C}[1]{>{\centering\arraybackslash}p{#1}}
\newcommand{\cmark}{\ding{51}}%
\newcommand{\xmark}{\ding{55}}%

\usepackage[breaklinks=true,bookmarks=false]{hyperref}

\cvprfinalcopy 

\def\cvprPaperID{****} 

\begin{document}

\title{MT-Net Submission to the Waymo 3D Detection Leaderboard}

\author{Shaoxiang Chen, Zequn Jie, Xiaolin Wei, Lin Ma\\
Meituan\\
}

\maketitle

\begin{abstract}
In this technical report, we introduce our submission to the Waymo 3D Detection leaderboard.
Our network is based on the Centerpoint architecture, but with significant improvements.
We design a 2D backbone to utilize multi-scale features for better detecting objects with various sizes, together with an optimal transport-based target assignment strategy, which dynamically assigns richer supervision signals to the detection candidates. 
We also apply test-time augmentation and model-ensemble for further improvements.
Our submission currently ranks 4th place with 78.45 mAPH on the Waymo 3D Detection leaderboard.
\end{abstract}

\section{Introduction}
3D object detection is indispensable in various real-world applications such as autonomous driving, and is receiving increasing attention from both the academia and industry in recent years.
The large-scale Waymo Open Dataset~\cite{sun2020scalability} serves as an important benchmark for 3D detection in autonomous driving and has enabled lots of new research. We are interested in Lidar-only 3D object detection. 
The Waymo Open Dataset provides high-quality data collected from Lidar sensors: 230k frames from 2.3k different driving scenes, and each frame has 177k points on average, and there are a total of 12M annotated 3D bounding boxes. 
We will introduce the method and experimental results of our submission in the following sections.

\section{Method}
In this section, we present the details of our 3D object detection method. We adopt the Centerpoint~\cite{yin2021center} architecture as a strong baseline, and make significant improvements to both the backbone and detection head.  
\subsection{Network}
\begin{figure}
	\centering
	\includegraphics[width=.95\linewidth]{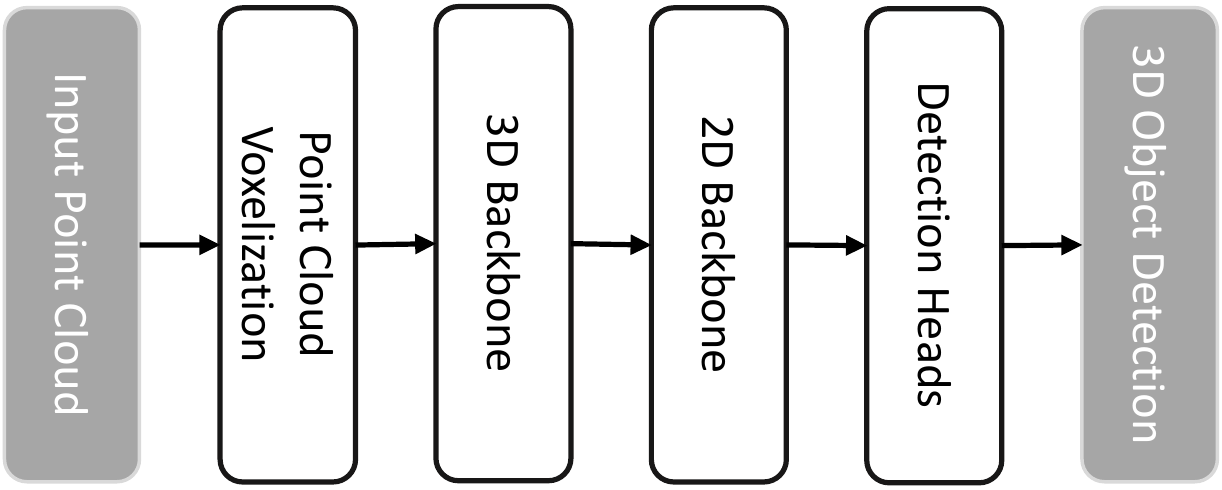}
	\caption{The general workflow of a 3D object detection method.}
	\label{fig1}
\end{figure}

Figure~\ref{fig1} demonstrates the general workflow of a 3D object detection method, we will introduce the detail of each part in the following sub-sections.
\subsubsection{Point Cloud Voxelization.} The input to our 3D object detection network is a point cloud, where each point has its features, including coordinates, reflectance, etc. We first voxelize the points (across x, y, z dimensions) within a fixed range using a fixed voxel size, and the point features within the same voxel are averaged as the voxel feature. The voxelized 3D input is then fed to the 3D feature extractor.

\subsubsection{3D Backbone.} The 3D feature extractor is composed of a series of 3D convolutional blocks. Each block has several 3D submanifold sparse convolutional layers~\cite{graham2017submanifold} and a final sparse convolutional layer, which downsamples the input by 2x along the z dimension and optionally downsamples 2x along the x and y dimensions. 
The total downsampling ratio of the 3D backbone is 4x. 
Each convolutional layer is followed by a batch-normalization layer and ReLU activation.

\subsubsection{2D Backbone.} The final output of the 3D feature extractor will be collapsed along the z dimension, resulting in a 2D feature map (pseudo image) and is further passed to the 2D feature extractor (backbone), which is responsible for extracting and combining multi-scale BEV features. 

We adopt two types of 2D backbones. 
The first one (abbreviated as 2D-A) is composed of self-calibrated convolutions (SC-Conv) blocks~\cite{liu2020improving} as in~\cite{ge2021real}. 
It directly combines multi-scale features maps upsampled from different blocks. The concatenated feature map has the same resolution as the input and is passed to the detection heads.
We also use another 2D backbone with FPN structure (abbreviated as 2D-B). It first down-samples the input via strided convolution in each block, then gradually up-samples the lower-resolution feature maps and fuses them with higher-resolution feature maps. The FPN features of different resolutions can be used for predicting objects of different sizes, e.g., vehicle and pedestrian.

\subsubsection{Predictions and Target Assignment} Following Centerpoint, the detection head of our network has 6 sub-heads for predicting the class heatmap, xy-offset, z-coordinate, 3D object size, orientation, and iou, respectively. 
But Centerpoint only assigns ground-truth labels to the corresponding center locations, and other surrounding locations' predictions are not supervised. 

We improve the target assignment of the sub-heads to obtain richer supervision signals for more accurate offset and size prediction. 
Specifically, we adopt the optimal transport assignment strategy~\cite{ge2021ota} to dynamically assign labels to more locations during training. 
Given a set ground-truth objects $\{\bm{g}_i\}_{i=1}^{N}$ and a set of current predictions $\{\bm{o}_j\}_{j=1}^{M}$,
we compute a pairwise cost matrix between the ground-truth objects and the predictions as 
\begin{equation*}
    C_{ij} = \mathcal{L}_{cls}(\bm{g}_i, \bm{o}_j) + \mathcal{L}_{reg}(\bm{g}_i, \bm{o}_j),
\end{equation*}
where $\mathcal{L}_{cls}$ is the BCE loss function, and $\mathcal{L}_{reg}$ is the $L_1$ loss.
We dynamically constrain that each ground-truth object can be assigned to $k_i$ predictions, where $k_i$ is the budget computed by summing the IoU values between it and all predictions. Then each prediction $\bm{o}_j$ gets the ground-truth label $\bm{g}_i$ with remaining budget and the minimum cost $C_{ij}$.

The final object prediction is generated by first taking the locations whose maximum class probabilities are over a threshold, and then adding the xy-offset and z-coordinate to produce 3D coordinates.

\subsection{Further Improvements}
We apply several techniques at the inference stage to further improve the final predictions. 

\textbf{Test-Time Augmentation.} We augment the input point cloud at test-time to obtain multiple versions of outputs. The augmentations applied are yaw-rotation with angles $[0.0,-0.13\pi,-0.07\pi,0.07\pi]$, global scaling with factors $[0.95, 1.05]$, and z-axis translation with offsets $[-0.2,0.2]$.

\textbf{Weighted Box Fusion.} Directly combining all the predictions from test-time augmentation and performing NMS could result in sub-optimal results. Since NMS discards the boxes that significantly overlap with higher confidence boxes, their information is not preserved. We adopt 3D weighted box fusion to process predicted boxes of each class and merge box locations and dimensions of matched boxes. The maximum number of resulting boxes is set to 500.

\textbf{GT-Paste Fading.} During training, we perform the GT-paste pre-processing, which directly copy some ground-truth object from a pre-built object database and paste them into the current frame. While this enriches the objects in each frame to let the model see more examples, the data distribution could drift away from the real distribution and hurt testing performance. To mediate this, we turn off the GT-paste augmentation at the last 5 epochs of training to make the model learn the real data distribution.

\textbf{Multiple Model Fusion.} In our final submission, we use two types of 2D backbones, and the models have different performances for each class, e.g., one of them may perform better for larger objects like vehicle and the other one performs better for cyclist. So we fuse predictions from multiple models before performing weighted box fusion according to their performances for each class.

\begin{table}[t]
\centering
\begin{tabular}{lc}
\hline
     Method & mAPH \\
     \hline
MPPNetEns-MMLab & 79.14\\
LIVOX\_Detection & 78.96\\
MT3D & 78.73\\
MT-Net (ours) & 78.45 \\
DeepFusion-Ens & 78.41\\
InceptioLidar & 77.84\\
AFDetV2-Ens & 77.64\\
Octopus\_Noah & 77.27\\
INT\_ensemble & 77.21\\
HorizonLiDAR3D &  77.11\\

\end{tabular}

\caption{The Waymo 3D detection Leaderboard, retrieved on July 11, 2022.}
\label{table:leaderboard}
\end{table}

\begin{table*}[t]

\centering

\begin{tabular}{ccc c ccc ccl}
     \#&Baseline & Label Assign & WBF \& TTA & +Val Data & 2D-B & GT-Paste Fading & mAPH \\
     \hline
     \hline
     1&\cmark & \xmark & \xmark & \xmark & \xmark & \xmark& 71.31 \\
     2&\cmark & \cmark & \xmark & \xmark & \xmark & \xmark& 71.91 \\
     3&\cmark & \cmark & \cmark & \xmark & \xmark & \xmark& 74.98 \\
     4&\cmark & \cmark & \cmark & \cmark & \xmark & \xmark& 76.16 \\
     5&\cmark & \cmark & \cmark & \cmark & \cmark & \xmark& 78.61 \\
     6&\cmark & \cmark & \cmark & \cmark & \cmark & \cmark& 80.13 \\
     \hline
\end{tabular}

\caption{Ablation studies on the Waymo val set (4k frames). The baseline is Centerpoint++ ~\cite{yin2021centerplus}, and we ablate each component of our method for comparison.}
\label{table:abl}
\end{table*}
\section{Experiments}

\subsection{Experiment Settings}
The Waymo Open Dataset has 158k frames for training (train set) and 40k frames for validation, we re-split these frames into 194k frames for training (trainval set) and 4k frames for validation (val set). We mainly test our models on the val set. Note that the trainval set includes samples from the original validation set, but any two frames of trainval set and val set are not from the same sequence. There are also 30k frames (test set) for submitting testing results to the leaderboard.

For each input frame, the points within range $[-75.2m, 75.2m]$, $[-75.2m, 75.2m]$, and $[-2m, 4m]$ in the x, y, and z axis are voxelized, the voxel size along x, y, and z axis are $0.1m$, $0.1m$ and $0.15m$, respectively. We also transform and fuse the points from previous two frames to densify the current point cloud. 
For data augmentation during training, we apply random flipping
along the x or y axis, global scaling with a factor sampled from $[0.95,1.05]$, random global rotation between $[-\pi/4,\pi/4]$, and random translation between $[-0.5m,0.5m]$.

We train the models using Pytorch, and use the AdamW optimizer with the one-cycle learning rate schedule. The max learning rate is 0.003, weight decay is 0.01, and momentum is 0.85 to 0.95. 
The training batch size is 16 and the samples are evenly distributed across 8 A100 GPUs.
We train the model for a total of 20 epochs. Training one model takes about 3 days.

\subsection{Results}
\textbf{Main Results.} Table~\ref{table:leaderboard} shows the entries of the Waymo 3D detection leaderboard as of July 11, 2022. Our submission achieves 78.45 mAPH, which is among the top-performing methods.

\textbf{Ablation Studies.}
Table~\ref{table:abl} shows the improvements brought by each component in our method. The baseline (\#1) is reproduced from the Centerpoint++ submission~\cite{yin2021centerplus}. Adding the optimal transport-based label assignment (\#2) brings 0.6 mAPH improvements. Further applying WBF (\#3) and TTA (\#4) to only a single model brings significant improvements (3 mAPH). Enlarging the training set with validation data (\#4) can also stably improve performance. The 2D backbone is also proved to be effective (\#5) and have a significant advantage over the baseline's 2D backbone (2.5 mAPH).
In addition, the GT-paste fading is also useful for both backbones, and we finally achieved 80.13 mAPH on the val set (\#6).

\section{Conclusion}

In this report, we present our 3D detection method with several techniques to improve performance. 
We adapted optimal transport-based label assignment to enrich the supervision signal and designed a 2D backbone to obtain multi-scale features for detecting objects of varying sizes. 
We also designed several simple yet effective strategies during training and inference to further improve 3D detection performance. 
Significant improvements over the baseline have been achieved and the final result on the Waymo 3D detection test set is 78.45 mAPH.

{\small
\bibliographystyle{ieee}

}

\end{document}